\documentclass[letterpaper, 10 pt, conference]{ieeeconf}  

\IEEEoverridecommandlockouts                              

\usepackage{amsmath} 
\usepackage{multirow}
\usepackage[font=small,skip=1pt]{caption} 
\usepackage{placeins} 
\usepackage{fontawesome5}
\usepackage{hyperref}
\usepackage{balance}
\usepackage[colorinlistoftodos]{todonotes}
\title{\LARGE \bf
AnchorVLA4D: an Anchor-Based Spatial-Temporal Vision-Language-Action Model for Robotic Manipulation
}
\author{Juan Zhu$^{1}$ 
Zhanying Shao$^{1}$ 
Xiaoqi Li$^{1,2}$ 
Ethan Morgan$^{1}$ 
Jiadong Xu$^{2}$ 
Hongwei Fan$^{1,2}$ 
Hao Dong$^{1,2,\ast}$
\thanks{$^{1}$PrimeBot}
\thanks{$^{2}$School of Computer Science, Peking University}
\thanks{$\ast$ Correspondence to \tt\small zhujuan.eileen@gmail.com}
}

\begin{document}

\maketitle
\thispagestyle{empty}
\pagestyle{empty}

\begin{abstract}

Since current Vision–Language–Action (VLA) systems suffer from limited spatial perception and the absence of memory throughout manipulation, we investigate visual anchors as a means to enhance spatial and temporal reasoning within VLA policies for robotic manipulation.

Conventional VLAs generate actions by conditioning on a single current frame together with a language instruction. However, since the frame is encoded as a 2D image, it does not contain detailed spatial information, and the VLA similarly lacks any means to incorporate past context. As a result, it frequently forgets objects under occlusion and becomes spatially disoriented during the manipulation process. Thus, we propose AnchorVLA4D, a simple spatial-temporal VLA that augments the visual input with an \emph{anchor} image to preserve the initial scene context throughout execution, and adds a lightweight spatial encoder that jointly processes the anchor and current frames to expose geometric relationships within an episode. 
Built on a Qwen2.5-VL backbone with a diffusion-based action head, AnchorVLA4D requires no additional sensing modalities (e.g., depth or point clouds) and introduces negligible inference overhead. 
Combining anchoring with a frozen pretrained spatial encoder yields further gains, realizing a 13.6\% improvement on the Simpler WidowX benchmark and confirming the approach on real-world tasks, where it achieved an average success rate of 80\%.

\end{abstract}

\section{INTRODUCTION}

Recently, Vision-Language-Action (VLA)\cite{kim2024openvla}\cite{black2026pi0visionlanguageactionflowmodel}\cite{li2024cogact} models have attracted significant attention due to their strong potential, which stems from leveraging the image–text understanding capabilities of Vision-Language Models (VLMs)\cite{yang2025qwen3}\cite{team2023gemini} trained on large-scale data. By attaching an action head to a VLM, a VLA system can generate appropriate action sequences to guide subsequent movements. Because robotic manipulation inherently depends on interactions governed by the state of the environment—which can often be represented by a single image or a sequence of images—using visual inputs from cameras in combination with natural language instructions from humans has proven to be an effective strategy.

\begin{figure}[thpb]
\centering
\includegraphics[width=\linewidth, keepaspectratio]{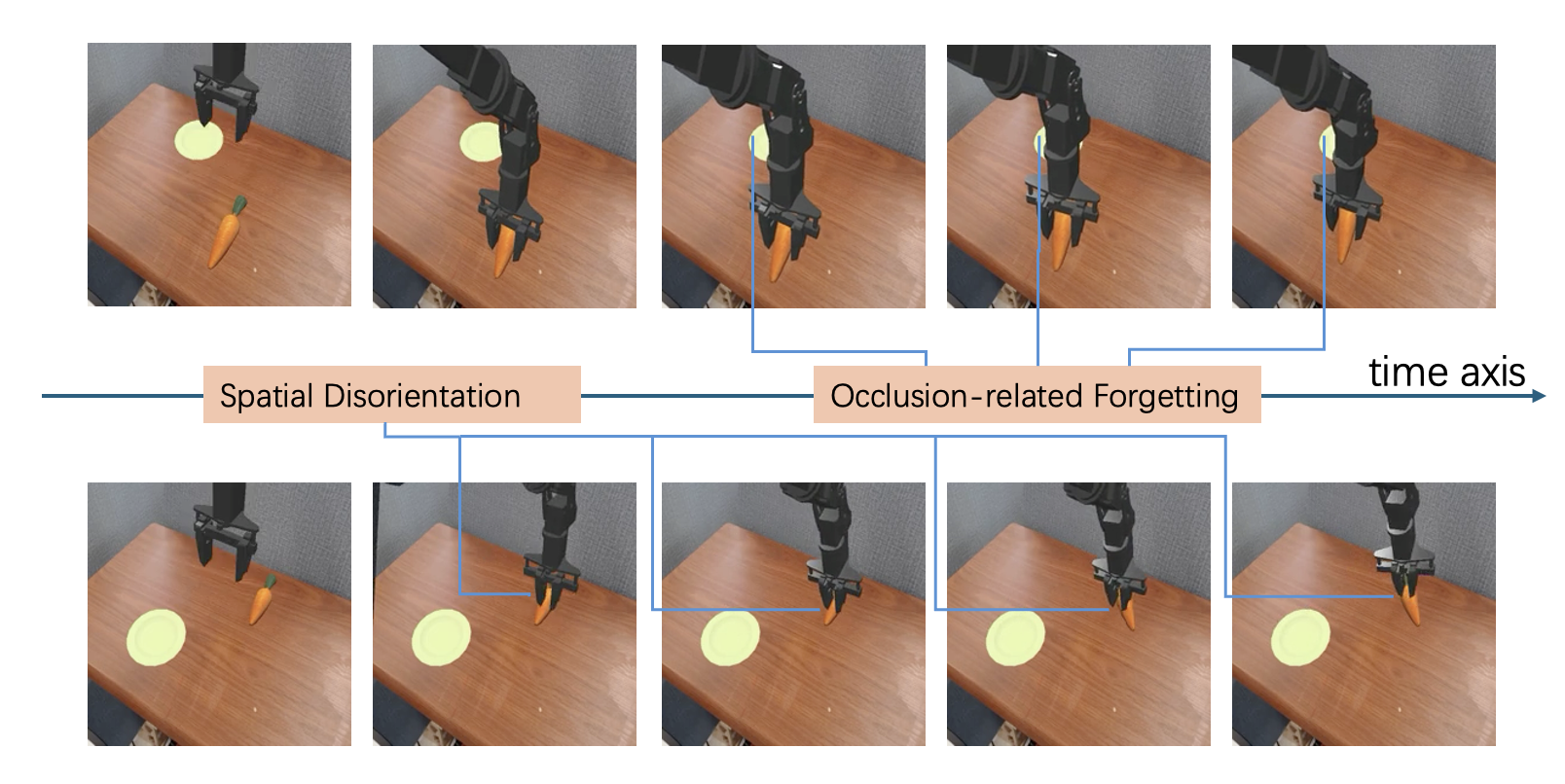}
\caption{Limits of depending on a single frame in conventional VLAs. The top row illustrates forgetting caused by occlusion, while the bottom row demonstrates spatial disorientation.}
\label{figure1}
\end{figure}

While VLAs have shown promising performance by exploiting the strong understanding capabilities of VLMs, the standard VLAs follow the prototype of \cite{kim2024openvla} and \cite{black2026pi0visionlanguageactionflowmodel}, where the current frame is combined with language instructions as input to a VLM, and the latent representation of the VLM is then connected to a downstream action expert, which can adopt a DiT architecture as in \cite{wen2025dexvla}. This design is relatively straightforward and can be scaled by leveraging the most prevalent types of datasets. Both training cost and inference latency remain acceptable to a certain degree. However, as illustrated in Figure~\ref{figure1}, three clear issues arise:
\begin{itemize}
    \item The model has only a limited understanding of past context; when the object is temporarily occluded by an action, the robot becomes confused and hesitates. As illustrated in the first row of Figure~\ref{figure1}, in the \textit{put carrot on plate} task, once the gripper grasped the carrot, the arm blocked the view of the target object—the plate—so the widow arm remained in the grasping state, unable to determine where to move next.
    This issue becomes more severe when observations are limited to a single viewpoint. Without additional perspectives, occlusions can leave the model functionally unable to see, making it difficult to complete the tasks.
    \item Because of insufficient spatial awareness, the VLA model cannot accurately grasp the object, which directly leads to the failure of the entire task. As shown in the second row of the figure, in one episode the arm attempted to grasp the carrot four times, but each attempt failed near the top of the object, suggesting some spatial disorientation, and it repeatedly made the same error without ever producing a successful trial.
    Although other modalities like depth images and point clouds are available and have been widely used to improve spatial understanding, they are typically expensive and not readily accessible in everyday environments. This restricts the range of applicable scenarios, limits the diversity and generality of the data, and ultimately prevents robots from acquiring broadly generalizable behaviors across different situations.
\end{itemize}

Existing methods continue to struggle with spatial reasoning and often suffer from memory forgetting. Our approach, AnchorVLA4D, is designed to overcome these limitations by relying on an additional visual cue.
By adding specific frames from each episode, we introduced only additional pre-existing visual inputs in order to examine the extent of potential performance gains.
Intuitively, we treat the first frame as an anchor for the entire action sequence, allowing the model to preserve a representation of the initial scene even when objects later become occluded by the manipulator arm. As expected, this anchoring causes the model to stop sooner when a failure occurs and to initiate a retry earlier. This leads to more frequent retries in failure scenarios and ultimately contributes to an overall increase in the success rate.

We then add a spatial encoder to jointly process the anchor frame and the current frame, as they provide spatial cues within a continuous episode.
The spatial encoder is able to process continuous frames and construct a 3D understanding of the scene. Since two frames already provide sufficient information for our task, while using too many frames would greatly increase computational cost, we limit the input accordingly. We observe that, after introducing the spatial encoder, retry accuracy improves and the system avoids failing at the same locations, effectively breaking the failure loop.
Our approach is effective, improving the final success rate in both simulation benchmarks and real-world evaluations. The implementation can be found at \href{https://github.com/eileenzhujuan/AnchorVLA4D}{AnchorVLA4D}.

Our contributions can be summarized as follows:
\begin{itemize}
    \item We present a vision-language-action model that achieves 64.6\% performance on the Simpler WidowX benchmark and an 80\% success rate on generalized real-world tasks.
    \item We leverage anchor frames to enhance the model's performance by 9.4\%, while adding only a minimal 8\% increase in inference latency.
    \item We demonstrate that integrating a pretrained spatial encoder provides effective guidance for more accurate spatial understanding, even without supervised labels, leading to a 4.2\% improvement at the cost of an additional 8\% latency.
\end{itemize}

\section{RELATED WORKS}

Vision-Language-Action (VLA) models have emerged as a promising paradigm for generalist robot control, building upon the rich visual and linguistic understanding of pretrained Vision-Language Models (VLMs). By attaching an action head to a VLM, VLAs can directly map visual observations and language instructions to robot actions. Representative works such as OpenVLA \cite{kim2024openvla}, $\pi_0$ \cite{black2026pi0visionlanguageactionflowmodel}, $\pi_{0.5}$ \cite{intelligence2025pi_}, and CogAct \cite{li2024cogact} have demonstrated strong generalization capabilities across diverse manipulation tasks. RoboFlamingo \cite{li2024roboflamingo} showed that pretrained VLMs can serve as effective robot imitators with minimal fine-tuning. More recent efforts have focused on scaling model capacity and improving action representations: Octo \cite{team2024octo} introduced an open-source generalist policy trained on 800k trajectories, RDT-1B \cite{liu2024rdt} scaled diffusion-based policies to 1.2 billion parameters for bimanual manipulation, and GR00T N1 \cite{nvidia2025groot} proposed a dual-system architecture for humanoid control. These advances establish the VLA architecture as a scalable foundation for robot learning.

To address the limited spatial perception inherent in standard VLA models, a number of prior works have explored incorporating explicit 3D spatial encoders. For example, 3D-VLA \cite{zhen20243dvla3dvisionlanguageactiongenerative} used point clouds to supervise an ideal target state. Along similar lines, SpatialVLA \cite{qu2025spatialvla} and PointVLA \cite{li2025pointvlainjecting3dworld} pretrained on 2D images and then incorporated 3D point cloud data to support fine-tuning. 3DS-VLA \cite{li20253ds} fused 2D images with point clouds to enhance spatial reasoning, while GeoVLA \cite{sun2025geovla} leveraged depth maps together with a point cloud network to model geometric features. DepthVLA \cite{yuan2025depthvla} and QDepthVLA \cite{li2025qdepth} introduced depth encoders to improve depth understanding. More recently, Evo-0 \cite{lin2025evo0} proposed a plug-and-play module that implicitly injects 3D geometry features, Avi \cite{huang2025avi} reframed action generation as volumetric 3D inference, GraphCoT-VLA \cite{chen2025graphcotvla} leveraged 3D pose-object graphs with chain-of-thought reasoning, RoboSplat \cite{li2025robosplat} employed 3D Gaussian splatting for demonstration generation. Moreover, FSD \cite{yuan2025seeingdoingbridgingreasoning} produces intermediate representations by reasoning over spatial relationships, thereby offering fine-grained guidance for robotic manipulation. VIPA-VLA \cite{zhang2025spatialaware} incorporates visual-physical alignment into the pretraining phase, while SoFAR \cite{qi2025sofar} builds a dataset of 3D objects labeled with semantic orientations to enhance spatial understanding. Despite these advances, most existing methods either require additional sensors or introduce significant computational overhead, limiting their practicality in real-world scenarios.

Beyond spatial enhancements, some studies attempt to exploit earlier frames to enhance model performance through temporal context. MemoryVLA \cite{shi2025memoryvla} employs a memory bank to capture the accumulation of historical information, while ST-VLA \cite{patratskiy2025spatial} incorporates the previous 30 frames as auxiliary inputs. TraceVLA \cite{zheng2024tracevla} derives a visual trace through a predefined procedure, which then serves as a manipulation prompt. More recent works have explored diverse temporal modeling strategies: ECoT \cite{zawalski2025ecot} introduces embodied chain-of-thought reasoning for multi-step planning, CoT-VLA \cite{zhao2025cotvla} predicts future image frames as visual reasoning intermediates, ACoT-VLA \cite{li2025acot} formulates reasoning as coarse action intent sequences, Long-VLA \cite{chen2025longvla} proposes phase-aware input masking for long-horizon tasks, MAP-VLA \cite{wang2025mapvla} augments action generation with demonstration-derived memory prompts, and EchoVLA \cite{liu2025echovla} incorporates declarative scene and episodic memory for mobile manipulation. However, approaches relying on extensive historical frames impose a substantial computational burden, motivating the need for more efficient temporal modeling strategies.

In contrast to these methods, we propose AnchorVLA4D, which leverages a single anchor frame (the first frame of each episode) combined with a lightweight spatial encoder to achieve efficient spatial-temporal modeling. This design preserves initial scene context without requiring multiple historical frames or additional sensing modalities, offering a practical solution for real-world deployment.

\section{METHODOLOGY}

\subsection{Preliminaries}
Given an episode composed of a sequence of visual observations ${I_0, I_1, \dots, I_i}$ and matching proprioceptive states ${S_0, S_1, \dots, S_i}$, along with a textual instruction $T$, the goal is to generate an action $a_i = [\Delta x_i, \Delta \theta_i, Gripper_i]$ for a single-arm manipulation scenario. Here, $x$ and $\theta$ denote the position and rotation, respectively, while $Gripper$ indicates the state of the robot gripper (open/close). Traditional VLA methods \cite{kim2024openvla}\cite{black2026pi0visionlanguageactionflowmodel} typically use only the current frame $I_i$ as visual input, while other approaches \cite{patratskiy2025spatial}\cite{shi2025memoryvla} additionally incorporate a subset of past frames to enrich the model with implicit contextual information. 

In our approach, we instead select an anchor image $I_{anchor}$ (in particular, we set $I_{anchor} = I_0$) together with the current image $I_i$ as inputs. Denoting the Vision-Language processing module as $VL$, the spatial encoder as $SE$, and the action head as $AH$, the overall procedure can be formulated as:

\begin{equation}
\begin{aligned}
[a_i \dots a_{i+j}]
  &= AH\Bigl(
      VL(I_{anchor}, I_i, T), \\
  &\qquad SE(I_{anchor}, I_i),\, S_i
     \Bigr)
\end{aligned}
\end{equation}

\begin{figure*}[tbpb]
\centering
\includegraphics[width=\linewidth, keepaspectratio]{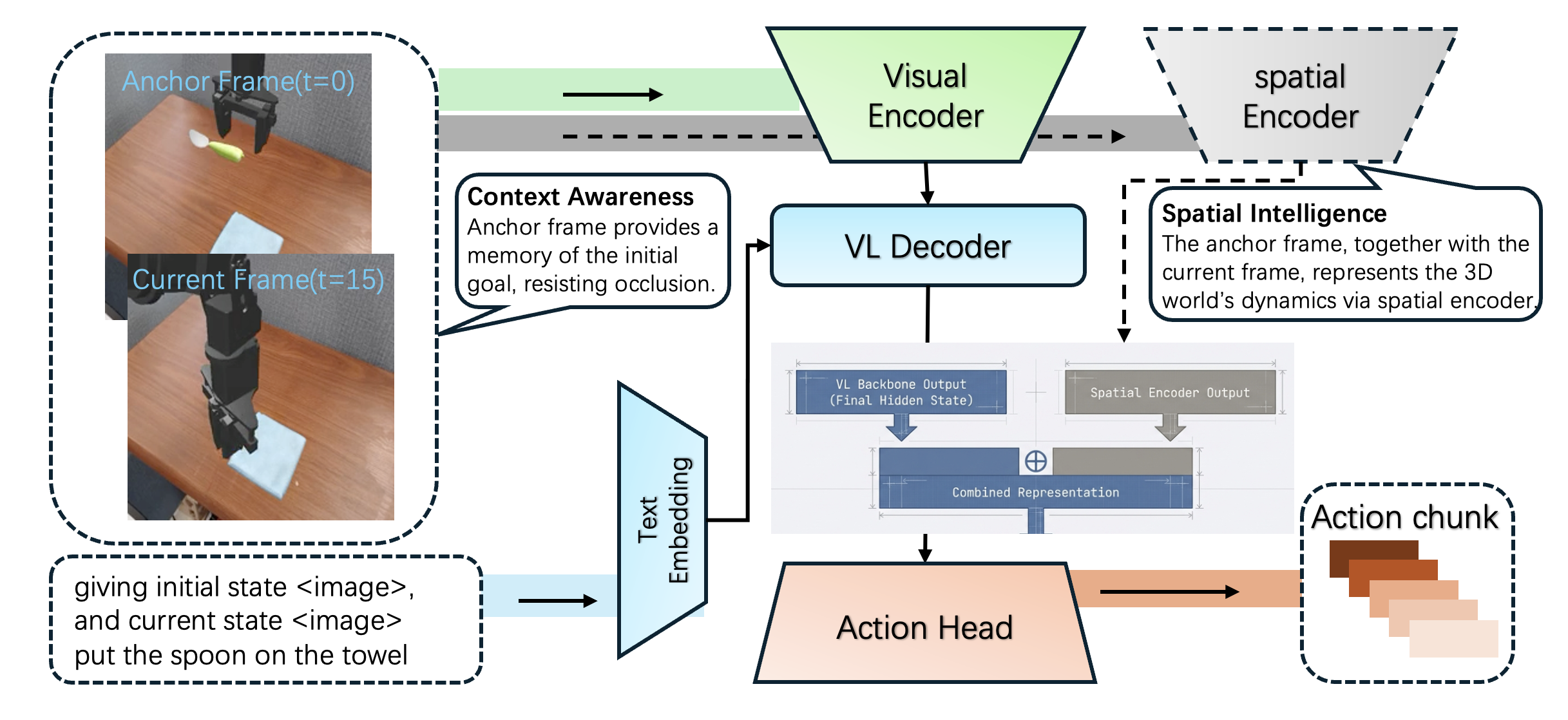}
\caption{Overall Architecture} 
\label{figure2}
\end{figure*}

\subsection{Model Architecture}
We employ Qwen2.5VL \cite{bai2025qwen25vltechnicalreport} 3B as the vision-language backbone and a 400M ScaleDP \cite{zhu2025scaling} model as the action head, both conditioned on the final hidden-layer representation and the proprioceptive state. The overall architecture is illustrated in Figure~\ref{figure2}: the VL decoder takes as input the outputs of the visual encoder and the text embeddings, and an action head is then used to generate the action chunks.

Owing to the versatility of Any4D \cite{karhade2025any4d}, we also conducted experiments where it served as the spatial encoder and was integrated into the hidden states of the VLM’s output.

We evaluate three model variants. The first is VanillaVLA, which uses neither anchors nor a spatial encoder. The second is AnchorVLA, which adds an anchor mechanism on top of VanillaVLA. The third is AnchorVLA4D, which combines both a spatial encoder and anchors.

\subsection{AnchorVLA4D Workflow}
We use the first frame of an episode as an anchor. This serves two main purposes: (1) it reminds the robot of the initial state while helping it filter out background noise, and (2) Along with the current observation, it explores a trajectory in the 3D physical environment, enabling the robot to build an understanding of the three-dimensional world.

The anchor frame and the current frame are fed into the network, which generates a sequence of hidden states; we take the final hidden state for subsequent processing. Additionally, if a spatial encoder is used, both images are also processed by this module. Because these frames show the same objects at different times, they help the model recall the original task goal and the objects’ motion trajectories. The spatial encoder’s output is then concatenated with the final hidden state from the VL backbone. The Action Head projects this combined representation into a fixed-length conditioning vector. Through a standard noise prediction procedure, we optimize the model by minimizing the MSE loss of the predicted noise. During inference, the model performs a denoising process, enabling it to reconstruct the correct action chunk from a noisy one.

\subsection{Training Recipe}
We train the model on an Ascend 910B node equipped with 8 NPUs, each with 64 GB of memory. The learning rate is set to 2e-5 and kept constant during pretraining on the full BridgeV2 dataset, while a cosine decay schedule is applied during finetuning on the 4 task-specific subsets. We use a weight decay of 1e-2. Pretraining is conducted for 30,000 steps with a global batch size of 512 per step, and finetuning is performed for 15,000 steps with a global batch size of 256 per step.

During the pretraining phase, all parameters, including those of the image encoder, are left unfrozen, since this component is crucial for transforming the model’s internal representation into the robotic manipulation domain. In the finetuning phase, we then selectively freeze certain parameters so that learning occurs only within specific parts of the model.

\section{EXPERIMENTS}

Our model utilizes weights loaded directly from the open-source Qwen2.5-VL\cite{bai2025qwen25vltechnicalreport} and the released ScaleDP-L\cite{wen2025dexvla} checkpoint.
\subsection{Simulation Experiments}
We train directly on the BridgeV2 \cite{walke2023bridgedata} datasets and subsequently evaluate the final success rate in the SimplerEnv \cite{li24simpler} environment using a single A800 GPU.
During training, we apply normalization to both the proprioceptive state and the action, following \cite{kim2024openvla}, and we jointly predict the next 5 actions as a single sequence, while at test time we execute only the first predicted action. This allows the model to leverage the context of future movements, improving its performance, consistent with the findings reported for ACT in \cite{zhao2023learning}. On average, each inference requires 0.215 seconds. The anchor and spatial encoder added 16\% latency relative to the original 0.185 seconds per inference, resulting in an effective control frequency of 4.65 Hz.

At test time, we evaluate 4 categories, each with 24 trials, and set the maximum episode length to 120 steps. Table~\ref{table1} reports the results.
\begin{table*}[h]
\caption{Simulation Success Rate}
\label{table1}
\centering
\setlength{\tabcolsep}{3pt} 
\renewcommand{\arraystretch}{1.1} 
\begin{tabular}{c|c|c|c|c|c|c|c}
\hline
\small Method Type & \small Model Size & \small Models & \small Spoon on Towel & \small Carrot on Plate & \small Stack Cube & \small Eggplant in Basket & \textbf{Average} \\
\hline
\multirow[c]{2}{*}{Standard} &
7B & OpenVLA\cite{kim2024openvla} & 4.2\% & 0.0\% & 0.0\% & 12.5\% & 4.2\% \\
& 7B + 300M & CogACT-Base\cite{li2024cogact} & 71.7\% & 50.8\% & 15.0\% & 67.5\% & 51.3\% \\
& 3B + 300M & $\pi_0$-Uniform$^{\star}$\cite{black2026pi0visionlanguageactionflowmodel} & 63.3\% & 58.8\% & 21.3\% & 79.2\% & 55.7\% \\
\hline

\multirow[c]{3}{*}{Spatial} &
3.5B & SpatialVLA\cite{qu2025spatialvla} & 16.7\% & 25.0\% & 29.2\% & \textbf{100\%} & 42.7\% \\
& 3.5B & SoFar\cite{qi2025sofar} & 58.3\% & 66.7\% & 70.8\% & 37.5\% & 58.3\%\\
& 13B & FSD\cite{yuan2025seeingdoingbridgingreasoning} & 41.7\% & 50.0\% & 33.3\% & 37.5\% & 40.6\% \\
\hline
\multirow[c]{3}{*}{Temporal} &
7B & TraceVLA\cite{zheng2024tracevla} & 12.5\% & 16.6\% & 16.6\% & 65.0\% & 27.7\% \\
& - & ST-VLA\cite{patratskiy2025spatial} & 36.4\% & 25.0\% & 36.4\% & 54.5\% & 38.1\%\\
& 7B + 300M & MemoryVLA\cite{shi2025memoryvla} & 75.0\% & \textbf{75.0\%} & 37.5\% & \textbf{100\%} & \textbf{71.9\%} \\ \hline

\multirow[c]{3}{*}{Ours} &
3B + 400M & VanillaVLA  & 54.2\% & 37.5\% & 33.3\% & 79.2\% & \underline{51.0\%} \\
& 3B + 400M & finetuned VanillaVLA & 75.0\% & 37.5\% & \textbf{62.5\%} & 70.8\% & 61.5\% \\
&4.4B & AnchorVLA4D$\star$ & \textbf{79.2\%} & 37.5\% & 50.0\% & 91.7\% & \textbf{64.6\%} \\ 
\hline
\end{tabular}
\end{table*} 

We divide existing methods into four categories. The first category is standard VLAs, which are trained without any additional spatial or temporal signals. This group includes OpenVLA, CogACT-Base, and $\pi_0$. The second category consists of VLAs that incorporate spatial information, such as SpatialVLA, SoFAR, and FSD. The third category includes temporally integrated VLAs, represented by TraceVLA, ST-VLA, and MemoryVLA. The final category is our own method: we begin by building a VanillaVLA without a spatial encoder and omitting anchors during training, and then, using this VanillaVLA as a baseline, we systematically investigate possible performance improvements.

While many models, such as \cite{kim2024openvla}\cite{black2026pi0visionlanguageactionflowmodel}\cite{shi2025memoryvla}, are pretrained on large-scale datasets like Open-XE\cite{o2024open}, our models are initialized from weights that have not been trained on any robotic data before. We present three variants of our approach. VanillaVLA is trained on the complete BridgeV2 dataset with 50 sampling timesteps, while the finetuned VanillaVLA starts from the pretrained VanillaVLA and is further trained on a subset of datasets using 50 sampling timesteps, which markedly increases latency compared to the other variants. The finetuned VanillaVLA demonstrates the feasibility of training from a single frame without requiring any additional data. To clearly investigate the benefits of the anchor mechanism, we conduct the following experiments using 10 sampling timesteps, which are both sufficient and efficient. Under this setting, AnchorVLA4D outperforms VanillaVLA by 13.6 percentage points, achieving a score of 64.6\%.

\begin{figure*}[tbhp]
\centering
\includegraphics[width=\linewidth, keepaspectratio]{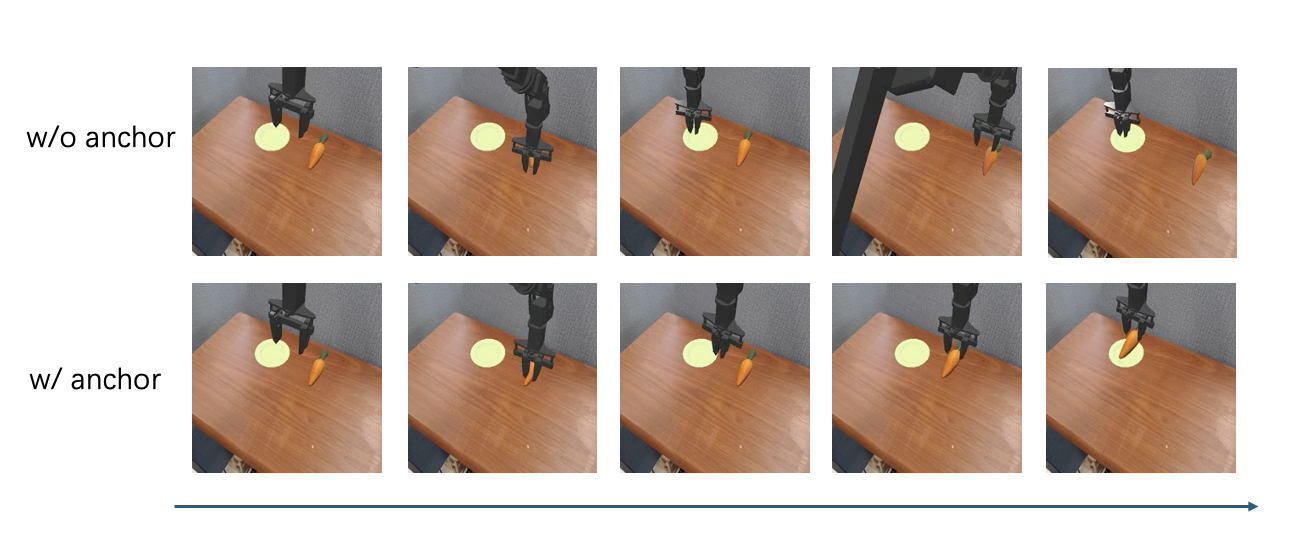}
\caption{More precise retries using an anchor}
\label{figure5}
\end{figure*}

Even though we do not perform large-scale pretraining on robotic data and our model is considerably smaller at 4.4B parameters, it still delivers strong performance. In contrast, MemoryVLA—the current state-of-the-art—relies on a 7.3B-parameter model whose backbone has been pretrained on a large-scale robot dataset. Moreover, our approach is both low-latency and straightforward. We believe this demonstrates that our method achieves genuinely strong performance. In addition, we have two main observations, as illustrated in Figure~\ref{figure5}:

\subsubsection{Early Retrying}
Our results indicate that adding an anchor clearly leads to retry attempts being initiated earlier. The anchored model tends to stop promptly upon encountering a failure and then restarts the process, whereas the vanilla version adheres to a predetermined trajectory and usually retries after completing the entire course.

\subsubsection{Improved Spatial Awareness}
After incorporating the anchor input, the model can better understand the environment and the spatial relationships between objects. To be more specific, given a previous failed attempt, the next attempt becomes more accurate. When an anchor is provided, the subsequent attempt succeeds, whereas without an anchor, the next attempt also fails.
This suggests that, in the subsequent trial, the model shows enhanced spatial understanding, resulting in a more precise grasp after the initial failure.

\subsection{Real World Experiments}
\begin{figure*}[tbhp]
\centering
\includegraphics[width=\linewidth, keepaspectratio]{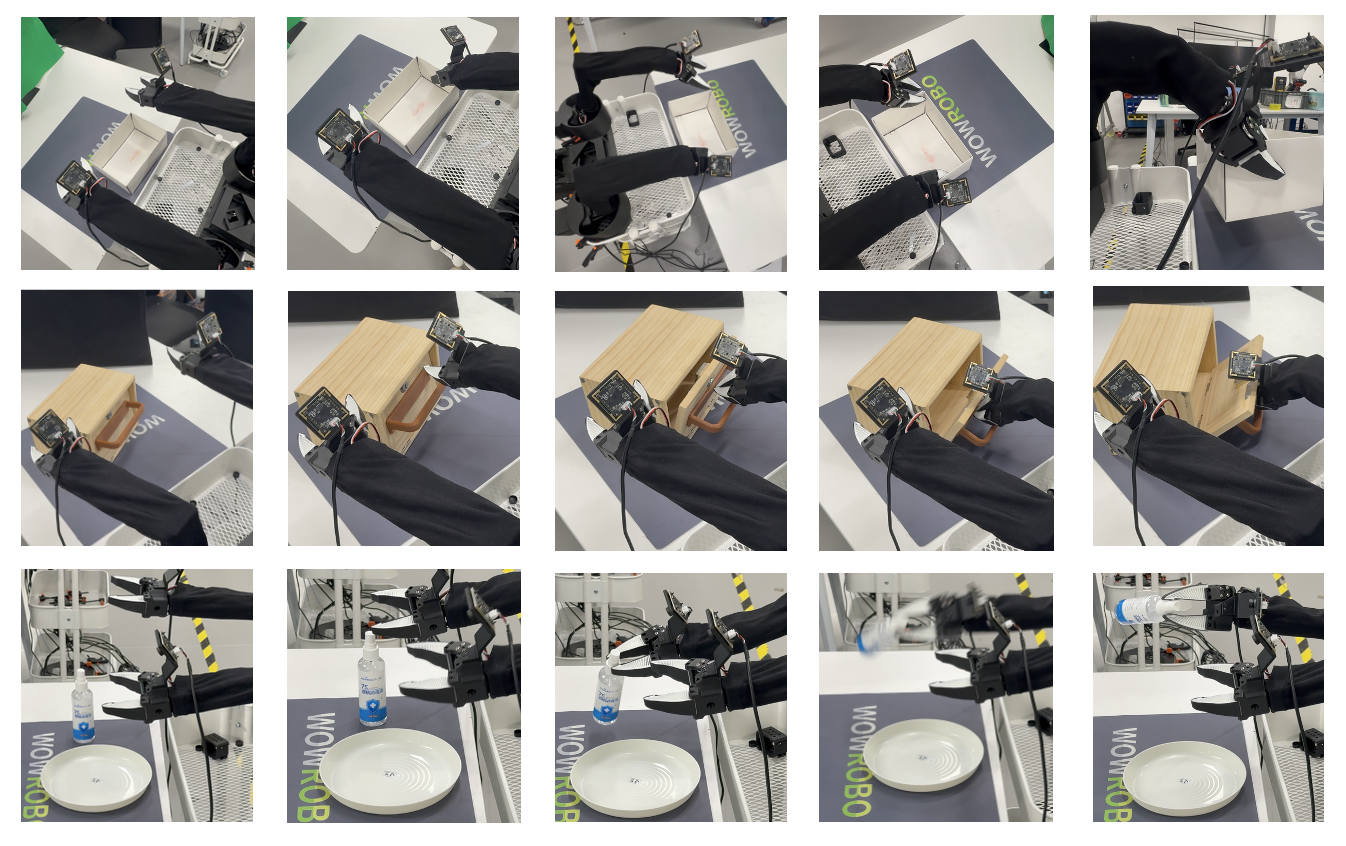}
\caption{Tasks in Real-World Environments}
\label{figure3}
\end{figure*}

We evaluate our model on xLerobot \cite{wang2025xlerobot}, an open-source, low-cost household dual-arm mobile robot that features a straightforward assembly process.
Our model is initialized from a trained checkpoint on the bridgeV2 dataset, and then further fine-tuned using our own collected data, with 30 episodes allocated to each task.
In total, we designed three tasks covering dual-arm and single-arm manipulation, a rotation task, and a hinge-joint task—as shown in Figure~\ref{figure3}.

\begin{itemize}
    \item Lift Box with Two Arms: This task requires the coordinated use of both arms, which is especially challenging for robots because the model has not previously encountered any examples of bimanual manipulation. The robot first grasps the right side of the box, then the left side, and finally lifts the box by moving both arms in unison.
    \item Open Drawer: In this task, we instruct the robot to open the drawer using its single arm. Since the drawer is a hinge-joint object, the arm must correctly infer the direction in which to apply force. In addition, when the arm approaches the drawer’s grasp point, occlusion becomes a significant challenge.
    \item Rotation: The robot arm needs to grasp a cup with water and then pour it onto a plate. This process requires a series of actions, including grasping, holding, and rotating.  
    
\end{itemize}

All inference runs are executed on an RTX 4090 Ti GPU. Each task is evaluated ten times. For xLerobot, we use three camera viewpoints, each with a resolution of 640x480, and include an additional anchor view, resulting in four images per inference. To lower both training and inference costs. We then resize each image to 320x240. The results indicate that this resolution is sufficient to capture the necessary information.
We find that the per-inference latency of AnchorVLA4D is around 300 ms. Compared with the simulation results, this increase is primarily caused by the higher image resolution, which goes from 224×224 in simulation to 320×240×4 in the real-world setup. To obtain smoother control, we run only the first 5 actions from each inference, even though each one generates 50 action chunks, resulting in an effective frequency of about 17 Hz.

\begin{figure}[tbhp]
\centering
\includegraphics[width=\linewidth, keepaspectratio]{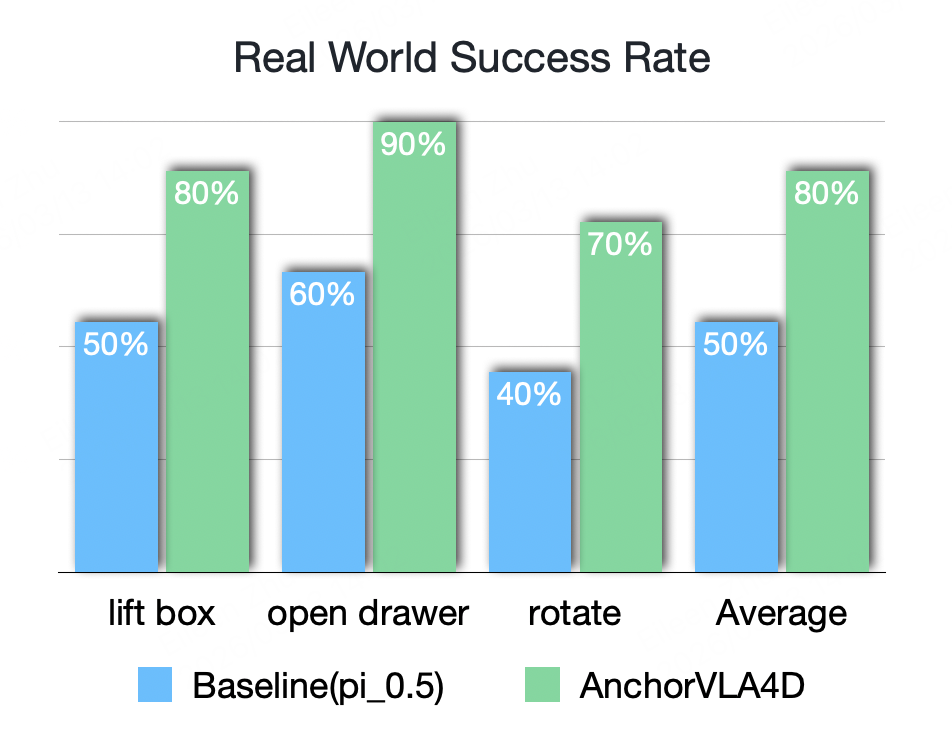}
\caption{Success rates for three different tasks}
\label{figure6}
\end{figure}

We take $\pi_{0.5}$ \cite{intelligence2025pi_} as our baseline since it has already been pretrained on multiple datasets, and we only further finetune it using data collected from xLerobot. During evaluation, we observed that using 50 chunks per run yields better performance than using 5 chunks, so all baseline results are reported using 50 action chunks for each inference. The overall results are presented in Figure~\ref{figure6} and depicted in Figure~\ref{figure3}.

We trained on BridgeV2 using 6D end-effector poses, whereas for XLerobot we relied on joint-space configurations. Moreover, the pre-training phase included only single-arm tasks, yet the resulting policy can successfully execute two-arm tasks after incorporating just 30 episodes of real-world data and 7,500 additional steps. This highlights the strong generalization capabilities of the model. Ultimately, we achieved an average success rate of 80\% across the three tasks, exceeding the 50\% baseline SR of $\pi_{0.5}$.

\section{ABLATION STUDIES}

\subsection{Performance Breakdown}
We break down the performance in Table~\ref{table3} and then discuss the performance improvements from two perspectives.
\subsubsection{Performance Gains from Adding Anchor}
After adding an anchor to VanillaVLA, the overall success rate on the SimplerEnv Widowx task increased by 9.4\%, indicating that the anchor serves as an informative input.

\subsubsection{Performance Gains from Adding Spatial Encoder}
By integrating a spatial encoder into AnchorVLA, we raise the overall success rate from 60.4\% to 64.6\%. This indicates that a pretrained spatial encoder can still be leveraged to enhance performance, even without explicit 3D supervision. Although Qwen2.5-VL already includes a ViT encoder—presumably capturing some degree of spatial structure—adding a dedicated spatial encoder nonetheless yields additional performance gains.

Since we do not have extra 3D supervision signals such as point clouds or depth maps, we explore two ways of incorporating the spatial encoder. In the first setting, we freeze all parameters of the spatial encoder and only finetune the immediately following layer to resolve the mismatch in embedding dimensions with the pretrained checkpoint. In the second setting, we unfreeze all parameters of the spatial encoder. As shown in Table~\ref{table3}, simply unfreezing the parameters causes the performance to drop by 6.2\% from its peak. An important observation is that training a separate spatial encoder from scratch yields almost no additional benefit in the absence of 3D supervised signals.

Moreover, we attempted to eliminate the spatial encoder by discarding its output after training. The test results show that overall performance drops by 6.3\%, indicating that the spatial encoder’s output indeed contributes to the final outcome.

\begin{table}[h]
\caption{Performance Breakdown}
\label{table3}
\centering
\setlength{\tabcolsep}{3pt} 
\renewcommand{\arraystretch}{1.1} 
\begin{tabular}{c|c|c|c|c}
\hline
\small Models & Anchor & SE & \small SR & Performance+- \\
\hline
VanillaVLA $\dagger$  & w/o & w/o & \underline{51.0\%} & - \\
AnchorVLA   & w/ & w/o & 60.4\% & $\dagger$ + 9.4\%\\
AnchorVLA4D $\ast$  & w/ & w/  & \textbf{64.6\%} & $\dagger$ + 13.6\%\\
\hline
AnchorVLA4D  & w/ & unfreeze & 59.4\% & $\ast$ - 6.2\% \\ 
AnchorVLA4D  & w/ & remove & 58.3\% & $\ast$ - 6.3\% \\ 
\hline
\end{tabular}
\end{table}

\subsection{Alternative Architectures and Design Choices}

We also experimented with other settings, such as using the previous 3 frames with a stride of 1 or 20. However, these configurations actually harmed the final performance. As shown in Table~\ref{table4}, we conducted an ablation study. For the Anchor format evaluation, we initialized the model with weights from VanillaVLA and continued training for 4k steps on a subset of datasets related to WidowX tasks. The resulting Success Rate (SR) clearly illustrates the performance difference. Using $I_0$ as the anchor yields an SR of 60.4\%, whereas using the past 3 frames reduces the SR to 46.9\%. Further increasing the stride between past frames to 20 causes a substantial drop in SR to 21.9\%.
We suspect this is because those frames introduced more noise than useful information.

\begin{table}[h]
\caption{Integration Ablation}
\label{table4}
\centering
\setlength{\tabcolsep}{3pt} 
\renewcommand{\arraystretch}{1.1} 
\begin{tabular}{c|c|c}
\hline
Anchor Format & SE Injection & \small SR\\
\hline
$I_0$ & - & \underline{55.2\%} \\
$I_{i-3 \dots i-1}$  & - & 46.9\% \\
$I_{i-60, i-40, i-20}$  & -  & 21.9\% \\
\hline
$I_0$  & Injection before Decoder & 0\% \\
$I_0$  & Injection in Action Head  & 46.9\%\\
$I_0$ & Concatenation & \underline{64.6\%} \\
\hline
\end{tabular}
\end{table}

Regarding the integration format between the spatial encoder output and the hidden states of the VL model, we experimented with three options: cross-attention before the text decoder, cross-attention inside the action expert, and concatenation before the action expert. Among these, the last approach performed best, as shown in Table~\ref{table4}. We hypothesize that the other two strategies could be more effective if trained with a more diverse dataset. However, given the limited data and a pretrained model that lacks any spatial-specific training or inputs, it is difficult for the model to learn all the necessary spatial representations and relationships. In contrast, concatenation, combined with an adaptive pooling layer, allows the model to flexibly fuse spatial information while preserving its original performance.

\subsection{Retry Statistics}
As shown in Table~\ref{table5}, Adding anchor leads to fewer overall retries because it achieves a higher success rate; however, in failure cases, the number of retries is slightly higher. Retrying remains important, as it offers a way to recover from being stuck.

\begin{table}[h]
\caption{Number of Retries}
\label{table5}
\centering
\setlength{\tabcolsep}{3pt} 
\renewcommand{\arraystretch}{1.1} 
\begin{tabular}{c|c|c|c}
\hline
Anchor & Overall Retries & in Case of Success & in Case of Failure \\
\hline
w/o & 1.54 & 1.43 & 1.66 \\
w/ & 1.40 & 1.21 & 1.68 \\
\hline
\end{tabular}
\end{table}

\subsection{Proprioceptive States}
Simulation training and testing offer much higher task variety and trajectory diversity than our real-world data collected using xLerobot. In simulation, proprioceptive states provide useful auxiliary information. However, in real-world environments, they hinder visual grounding ability. We found that removing proprioceptive state input improves visual grounding by eliminating the possibility of simply memorizing trajectories. As presented in Table~\ref{table6}, we conduct ablation studies on the proprioceptive state inputs. Under the same number of training steps (10k for simulation and 5k for the real-world open-drawer task), we find that, after introducing these states, the success rate in simulation increases, whereas the success rate in the real-world experiments decreases.

\begin{table}[h]
\caption{Comparison With or Without Proprioceptive States}
\label{table6}
\centering
\setlength{\tabcolsep}{3pt} 
\renewcommand{\arraystretch}{1.1} 
\begin{tabular}{c|c|c}
\hline
Proprioceptive States & Simulation(10k) & Realworld Open Drawer(5k) \\
\hline
w/o & 40.6\% & 60\% \\
w/ & 51.0\% & 50\% \\
\hline
\end{tabular}
\end{table}

\section{CONCLUSIONS}

By leveraging the first frame of an episode as an anchor to give the model an initial cue about the task and spatial configuration, our AnchorVLA4D approach significantly enhances overall performance with minimal extra overhead, achieving a 13.6\% improvement and reaching 64.6\% on the SimplerEnv benchmark, as well as an 80\% success rate in real-world experiments.
Constraining the temporal window to a fixed, short duration enables the model to keep its memory current while still maintaining access to earlier contextual information.

Nevertheless, the advantage of using an anchor comes with a corresponding drawback: when the execution state drifts too far from the initial state, our tasks become more likely to fail. This situation frequently arises when the gripper is vertically aligned with the objects in simulation tests. In such cases, the gripper must rotate to obtain a better grasp. Failures usually occur during the grasping phase, while the object remains stationary and the gripper must move a large distance from its original pose to reach the target pose. Consequently, the anchor can no longer provide meaningful guidance for retries or corrections and instead imposes a bias toward the original state, which increases the failure rate. This type of failure could be especially common in long-duration tasks, where the anchor becomes an outdated memory. A potential remedy is to incrementally update the anchor as the task progresses.

Building on the benefits and limitations of AnchorVLA4D, we argue that the capabilities of VLAs are fundamentally bounded by their underlying vision-language models. By adapting these models to robotic domains, our work establishes a foundation for exploring how rapidly growing robotic datasets can be leveraged to improve manipulation capabilities, aiming to address existing limitations in spatiotemporal reasoning. By identifying and leveraging subtle visual signals, we can significantly strengthen these capabilities. We aim for our study to encourage further exploration into more effective methods.

\section*{ACKNOWLEDGEMENT}
The experimental and computational work in this research run on the Huawei Cloud AI Compute Service. We appreciate the stable compute supply from this platform. Meanwhile, this work was partially supported by PKU Kunpeng \& Ascend Center of Excellence.


{
    \small
    \bibliographystyle{IEEEtran}
    \bibliography{ref}

@misc{zhen20243dvla3dvisionlanguageactiongenerative,
      title={3D-VLA: A 3D Vision-Language-Action Generative World Model}, 
      author={Haoyu Zhen and Xiaowen Qiu and Peihao Chen and Jincheng Yang and Xin Yan and Yilun Du and Yining Hong and Chuang Gan},
      year={2024},
      eprint={2403.09631},
      archivePrefix={arXiv},
      primaryClass={cs.CV},
      url={https://arxiv.org/abs/2403.09631}, 
}

@misc{li2025pointvlainjecting3dworld,
      title={PointVLA: Injecting the 3D World into Vision-Language-Action Models}, 
      author={Chengmeng Li and Junjie Wen and Yan Peng and Yaxin Peng and Feifei Feng and Yichen Zhu},
      year={2025},
      eprint={2503.07511},
      archivePrefix={arXiv},
      primaryClass={cs.RO},
      url={https://arxiv.org/abs/2503.07511}, 
}

@article{shi2025memoryvla,
  title={Memoryvla: Perceptual-cognitive memory in vision-language-action models for robotic manipulation},
  author={Shi, Hao and Xie, Bin and Liu, Yingfei and Sun, Lin and Liu, Fengrong and Wang, Tiancai and Zhou, Erjin and Fan, Haoqiang and Zhang, Xiangyu and Huang, Gao},
  journal={arXiv preprint arXiv:2508.19236},
  year={2025}
}

@article{patratskiy2025spatial,
  title={Spatial traces: Enhancing vla models with spatial-temporal understanding},
  author={Patratskiy, Maxim A and Kovalev, Alexey K and Panov, Aleksandr I},
  journal={Optical Memory and Neural Networks},
  volume={34},
  number={Suppl 1},
  pages={S72--S82},
  year={2025},
  publisher={Springer}
}

@article{qu2025spatialvla,
  title={Spatialvla: Exploring spatial representations for visual-language-action model},
  author={Qu, Delin and Song, Haoming and Chen, Qizhi and Yao, Yuanqi and Ye, Xinyi and Ding, Yan and Wang, Zhigang and Gu, JiaYuan and Zhao, Bin and Wang, Dong and others},
  journal={arXiv preprint arXiv:2501.15830},
  year={2025}
}

@inproceedings{li20253ds,
  title={3ds-vla: A 3d spatial-aware vision language action model for robust multi-task manipulation},
  author={Li, Xiaoqi and Heng, Liang and Liu, Jiaming and Shen, Yan and Gu, Chenyang and Liu, Zhuoyang and Chen, Hao and Han, Nuowei and Zhang, Renrui and Tang, Hao and others},
  booktitle={9th Annual Conference on Robot Learning},
  year={2025}
}

@article{sun2025geovla,
  title={Geovla: Empowering 3d representations in vision-language-action models},
  author={Sun, Lin and Xie, Bin and Liu, Yingfei and Shi, Hao and Wang, Tiancai and Cao, Jiale},
  journal={arXiv preprint arXiv:2508.09071},
  year={2025}
}

@misc{black2026pi0visionlanguageactionflowmodel,
      title={$\pi_0$: A Vision-Language-Action Flow Model for General Robot Control}, 
      author={Kevin Black and Noah Brown and Danny Driess and Adnan Esmail and Michael Equi and Chelsea Finn and Niccolo Fusai and Lachy Groom and Karol Hausman and Brian Ichter and Szymon Jakubczak and Tim Jones and Liyiming Ke and Sergey Levine and Adrian Li-Bell and Mohith Mothukuri and Suraj Nair and Karl Pertsch and Lucy Xiaoyang Shi and James Tanner and Quan Vuong and Anna Walling and Haohuan Wang and Ury Zhilinsky},
      year={2026},
      eprint={2410.24164},
      archivePrefix={arXiv},
      primaryClass={cs.LG},
      url={https://arxiv.org/abs/2410.24164}, 
}

@article{wen2025dexvla,
  title={Dexvla: Vision-language model with plug-in diffusion expert for general robot control},
  author={Wen, Junjie and Zhu, Yichen and Li, Jinming and Tang, Zhibin and Shen, Chaomin and Feng, Feifei},
  journal={arXiv preprint arXiv:2502.05855},
  year={2025}
}

@article{kim2024openvla,
  title={Openvla: An open-source vision-language-action model},
  author={Kim, Moo Jin and Pertsch, Karl and Karamcheti, Siddharth and Xiao, Ted and Balakrishna, Ashwin and Nair, Suraj and Rafailov, Rafael and Foster, Ethan and Lam, Grace and Sanketi, Pannag and others},
  journal={arXiv preprint arXiv:2406.09246},
  year={2024}
}

@article{karhade2025any4d,
  title={Any4D: Unified Feed-Forward Metric 4D Reconstruction},
  author={Karhade, Jay and Keetha, Nikhil and Zhang, Yuchen and Gupta, Tanisha and Sharma, Akash and Scherer, Sebastian and Ramanan, Deva},
  journal={arXiv preprint arXiv:2512.10935},
  year={2025}
}

@misc{bai2025qwen25vltechnicalreport,
      title={Qwen2.5-VL Technical Report}, 
      author={Shuai Bai and Keqin Chen and Xuejing Liu and Jialin Wang and Wenbin Ge and Sibo Song and Kai Dang and Peng Wang and Shijie Wang and Jun Tang and Humen Zhong and Yuanzhi Zhu and Mingkun Yang and Zhaohai Li and Jianqiang Wan and Pengfei Wang and Wei Ding and Zheren Fu and Yiheng Xu and Jiabo Ye and Xi Zhang and Tianbao Xie and Zesen Cheng and Hang Zhang and Zhibo Yang and Haiyang Xu and Junyang Lin},
      year={2025},
      eprint={2502.13923},
      archivePrefix={arXiv},
      primaryClass={cs.CV},
      url={https://arxiv.org/abs/2502.13923}, 
}

@inproceedings{zhu2025scaling,
  title={Scaling diffusion policy in transformer to 1 billion parameters for robotic manipulation},
  author={Zhu, Minjie and Zhu, Yichen and Li, Jinming and Wen, Junjie and Xu, Zhiyuan and Liu, Ning and Cheng, Ran and Shen, Chaomin and Peng, Yaxin and Feng, Feifei and others},
  booktitle={2025 IEEE International Conference on Robotics and Automation (ICRA)},
  pages={10838--10845},
  year={2025},
  organization={IEEE}
}

@inproceedings{walke2023bridgedata,
  title={Bridgedata v2: A dataset for robot learning at scale},
  author={Walke, Homer Rich and Black, Kevin and Zhao, Tony Z and Vuong, Quan and Zheng, Chongyi and Hansen-Estruch, Philippe and He, Andre Wang and Myers, Vivek and Kim, Moo Jin and Du, Max and others},
  booktitle={Conference on Robot Learning},
  pages={1723--1736},
  year={2023},
  organization={PMLR}
}

@article{li24simpler,
         title={Evaluating Real-World Robot Manipulation Policies in Simulation},
         author={Xuanlin Li and Kyle Hsu and Jiayuan Gu and Karl Pertsch and Oier Mees and Homer Rich Walke and Chuyuan Fu and Ishikaa Lunawat and Isabel Sieh and Sean Kirmani and Sergey Levine and Jiajun Wu and Chelsea Finn and Hao Su and Quan Vuong and Ted Xiao},
         journal = {arXiv preprint arXiv:2405.05941},
         year={2024}
}

@misc{wang2025xlerobot,
    author = {Wang, Gaotian and Lu, Zhuoyi and Yiyang, Huang and Yihao, Liu},
    title = {XLeRobot: A Practical Low-cost Household Dual-Arm Mobile Robot Design for General Manipulation},
    howpublished = "\url{https://github.com/Vector-Wangel/XLeRobot}",
    year = {2025}
}

@article{zhao2023learning,
  title={Learning fine-grained bimanual manipulation with low-cost hardware},
  author={Zhao, Tony Z and Kumar, Vikash and Levine, Sergey and Finn, Chelsea},
  journal={arXiv preprint arXiv:2304.13705},
  year={2023}
}

@article{yang2025qwen3,
  title={Qwen3 technical report},
  author={Yang, An and Li, Anfeng and Yang, Baosong and Zhang, Beichen and Hui, Binyuan and Zheng, Bo and Yu, Bowen and Gao, Chang and Huang, Chengen and Lv, Chenxu and others},
  journal={arXiv preprint arXiv:2505.09388},
  year={2025}
}

@article{team2023gemini,
  title={Gemini: a family of highly capable multimodal models},
  author={Team, Gemini and Anil, Rohan and Borgeaud, Sebastian and Alayrac, Jean-Baptiste and Yu, Jiahui and Soricut, Radu and Schalkwyk, Johan and Dai, Andrew M and Hauth, Anja and Millican, Katie and others},
  journal={arXiv preprint arXiv:2312.11805},
  year={2023}
}

@article{li2024cogact,
  title={Cogact: A foundational vision-language-action model for synergizing cognition and action in robotic manipulation},
  author={Li, Qixiu and Liang, Yaobo and Wang, Zeyu and Luo, Lin and Chen, Xi and Liao, Mozheng and Wei, Fangyun and Deng, Yu and Xu, Sicheng and Zhang, Yizhong and others},
  journal={arXiv preprint arXiv:2411.19650},
  year={2024}
}

@article{zheng2024tracevla,
  title={Tracevla: Visual trace prompting enhances spatial-temporal awareness for generalist robotic policies},
  author={Zheng, Ruijie and Liang, Yongyuan and Huang, Shuaiyi and Gao, Jianfeng and Daum{\'e} III, Hal and Kolobov, Andrey and Huang, Furong and Yang, Jianwei},
  journal={arXiv preprint arXiv:2412.10345},
  year={2024}
}

@article{qi2025sofar,
  title={Sofar: Language-grounded orientation bridges spatial reasoning and object manipulation},
  author={Qi, Zekun and Zhang, Wenyao and Ding, Yufei and Dong, Runpei and Yu, Xinqiang and Li, Jingwen and Xu, Lingyun and Li, Baoyu and He, Xialin and Fan, Guofan and others},
  journal={arXiv preprint arXiv:2502.13143},
  year={2025}
}

@article{li2025qdepth,
  title={QDepth-VLA: quantized depth prediction as auxiliary supervision for vision-language-action models},
  author={Li, Yixuan and Chen, Yuhui and Zhou, Mingcai and Li, Haoran and Zhang, Zhengtao and Zhao, Dongbin},
  journal={arXiv preprint arXiv:2510.14836},
  year={2025}
}

@article{yuan2025depthvla,
  title={Depthvla: Enhancing vision-language-action models with depth-aware spatial reasoning},
  author={Yuan, Tianyuan and Liu, Yicheng and Lu, Chenhao and Chen, Zhuoguang and Jiang, Tao and Zhao, Hang},
  journal={arXiv preprint arXiv:2510.13375},
  year={2025}
}

@inproceedings{o2024open,
  title={Open x-embodiment: Robotic learning datasets and rt-x models: Open x-embodiment collaboration 0},
  author={O’Neill, Abby and Rehman, Abdul and Maddukuri, Abhiram and Gupta, Abhishek and Padalkar, Abhishek and Lee, Abraham and Pooley, Acorn and Gupta, Agrim and Mandlekar, Ajay and Jain, Ajinkya and others},
  booktitle={2024 IEEE International Conference on Robotics and Automation (ICRA)},
  pages={6892--6903},
  year={2024},
  organization={IEEE}
}

@article{lin2025evo0,
  title={Evo-0: Vision-Language-Action Model with Implicit Spatial Understanding},
  author={Lin, Tao and Li, Gen and Zhong, Yilei and Zou, Yanwen and Zhao, Bo},
  journal={arXiv preprint arXiv:2507.00416},
  year={2025}
}

@article{team2024octo,
  title={Octo: An Open-Source Generalist Robot Policy},
  author={Team, Octo Model and Ghosh, Dibya and Walke, Homer and Pertsch, Karl and Black, Kevin and Mees, Oier and Dasari, Sudeep and Hejna, Joey and Kreber, Tobias and Finn, Chelsea and Levine, Sergey},
  journal={arXiv preprint arXiv:2405.12213},
  year={2024}
}

@article{liu2024rdt,
  title={RDT-1B: a Diffusion Foundation Model for Bimanual Manipulation},
  author={Liu, Songming and Wu, Lingxuan and Li, Bangguo and Tan, Hengkai and Chen, Huayu and Wang, Zhengyi and Xu, Ke and Su, Hang and Zhu, Jun},
  journal={arXiv preprint arXiv:2410.07864},
  year={2024}
}

@article{nvidia2025groot,
  title={GR00T N1: An Open Foundation Model for Generalist Humanoid Robots},
  author={NVIDIA and Bjorck, Johan and Bronars, Abigail and Byeon, Wonmin and Chao, Yu-Wei and Cheng, Yuke and Duan, Haoran and Dunn, Ethan and Fan, Linxi and Feng, Yuxiang and others},
  journal={arXiv preprint arXiv:2503.14734},
  year={2025}
}

@article{huang2025avi,
  title={Avi: A 3D Vision-Language-Action Model Architecture Generating Action from Volumetric Inference},
  author={Huang, Xiaoyu and Chen, Hao and Zhang, Yifan and Wang, He and Liu, Huaping},
  journal={arXiv preprint arXiv:2510.21746},
  year={2025}
}

@article{chen2025graphcotvla,
  title={GraphCoT-VLA: A 3D Spatial-Aware Reasoning Vision-Language-Action Model for Robotic Manipulation with Ambiguous Instructions},
  author={Chen, Zixuan and Wang, Yifan and Liu, Jiaming and Zhang, Renrui and Dong, Hao and Pang, Jianshu},
  journal={arXiv preprint arXiv:2508.07650},
  year={2025}
}

@article{li2025robosplat,
  title={RoboSplat: Novel Demonstration Generation with Gaussian Splatting Enables Robust One-Shot Manipulation},
  author={Li, Shuo and Zhang, Yifan and Chen, Hao and Wang, He},
  journal={arXiv preprint arXiv:2504.13175},
  year={2025}
}

@article{zhang2025spatialaware,
  title={Spatial-Aware VLA Pretraining through Visual-Physical Alignment from Human Videos},
  author={Zhang, Yuxiang and Chen, Hao and Liu, Jiaming and Wang, He and Liu, Huaping},
  journal={arXiv preprint arXiv:2512.13080},
  year={2025}
}

@inproceedings{zawalski2025ecot,
  title={Robotic Control via Embodied Chain-of-Thought Reasoning},
  author={Zawalski, Michal and Chen, William and Pertsch, Karl and Mees, Oier and Finn, Chelsea and Levine, Sergey},
  booktitle={9th Annual Conference on Robot Learning},
  year={2025}
}

@inproceedings{zhao2025cotvla,
  title={CoT-VLA: Visual Chain-of-Thought Reasoning for Vision-Language-Action Models},
  author={Zhao, Qingqing and Xu, Yilun and Pathak, Deepak and Gupta, Abhinav},
  booktitle={Proceedings of the IEEE/CVF Conference on Computer Vision and Pattern Recognition},
  year={2025}
}

@article{chen2025longvla,
  title={Long-VLA: Unleashing Long-Horizon Capability of Vision Language Action Model for Robot Manipulation},
  author={Chen, Yucheng and Wang, Tiancai and Liu, Xudong and Shi, Hao and Xie, Bin and Liu, Yingfei and Fan, Haoqiang and Zhang, Xiangyu},
  journal={arXiv preprint arXiv:2508.19958},
  year={2025}
}

@article{wang2025mapvla,
  title={MAP-VLA: Memory-Augmented Prompting for Vision-Language-Action Model in Robotic Manipulation},
  author={Wang, Yifan and Chen, Hao and Zhang, Yifan and Liu, Jiaming and Wang, He},
  journal={arXiv preprint arXiv:2511.09516},
  year={2025}
}

@article{liu2025echovla,
  title={EchoVLA: Robotic Vision-Language-Action Model with Synergistic Declarative Memory for Mobile Manipulation},
  author={Liu, Yuxiang and Chen, Hao and Zhang, Yifan and Wang, He and Liu, Huaping},
  journal={arXiv preprint arXiv:2511.18112},
  year={2025}
}

@article{li2025acot,
  title={ACoT-VLA: Action Chain-of-Thought for Vision-Language-Action Models},
  author={Li, Xinghang and Liu, Peiyan and Chen, Yixiang and Wu, Hongtao and Huang, Yan and Wang, Liang and Kong, Tao},
  journal={arXiv preprint arXiv:2601.11404},
  year={2025}
}

@inproceedings{li2024roboflamingo,
  title={Vision-Language Foundation Models as Effective Robot Imitators},
  author={Li, Xinghang and Liu, Minghuan and Zhang, Hanbo and Yu, Cunjun and Xu, Jie and Wu, Hongtao and Cheang, Chilam and Jing, Ya and Zhang, Weinan and Liu, Huaping and Li, Hang and Kong, Tao},
  booktitle={International Conference on Learning Representations},
  year={2024}
}

@misc{yuan2025seeingdoingbridgingreasoning,
      title={From Seeing to Doing: Bridging Reasoning and Decision for Robotic Manipulation}, 
      author={Yifu Yuan and Haiqin Cui and Yibin Chen and Zibin Dong and Fei Ni and Longxin Kou and Jinyi Liu and Pengyi Li and Yan Zheng and Jianye Hao},
      year={2025},
      eprint={2505.08548},
      archivePrefix={arXiv},
      primaryClass={cs.RO},
      url={https://arxiv.org/abs/2505.08548}, 
}

@article{intelligence2025pi_,
  title={$\pi_{0.5}$: a Vision-Language-Action Model with Open-World Generalization},
  author={Intelligence, Physical and Black, Kevin and Brown, Noah and Darpinian, James and Dhabalia, Karan and Driess, Danny and Esmail, Adnan and Equi, Michael and Finn, Chelsea and Fusai, Niccolo and others},
  journal={arXiv preprint arXiv:2504.16054},
  year={2025}
}
}

\end{document}